\documentclass[letterpaper, 10 pt, conference]{IEEEconf}

\usepackage[
    letterpaper,
    top=54pt,
    bottom=55pt,
    left=54pt,
    right=54pt
]{geometry}

\usepackage{amsmath,amsfonts}
\usepackage{algorithmic}
\usepackage{algorithm}
\usepackage{array}
\usepackage{textcomp}
\usepackage{stfloats}
\usepackage{url}
\usepackage{verbatim}
\usepackage{graphicx}
\usepackage{wrapfig}
\usepackage{cite}
\usepackage{balance}
\usepackage{stfloats}
\usepackage{booktabs}
\usepackage{color}
\usepackage[dvipsnames]{xcolor}
\usepackage{dsfont}
\usepackage{colortbl}
\usepackage{bm}
\definecolor{DBlue}{HTML}{1E90FF}
\usepackage{soul}
\usepackage[
    mode=buildnew
]{standalone}

\newif\ifshowhighlights
\showhighlightsfalse  

\newcommand{\changed}[1]{%
  \ifshowhighlights\textcolor{blue}{#1}\else#1\fi%
}

\usepackage{tikz}
\usepackage{tikz-3dplot}
\usetikzlibrary{positioning, shapes.geometric, arrows.meta, fit, calc}
\usepackage{ragged2e}

\usepackage{makecell}
\usepackage{booktabs}
\usepackage{amssymb}
\usepackage{bbding}
\usepackage{pifont}
\usepackage{wasysym}
\usepackage{subfigure}

\definecolor{companionlight}{RGB}{240,245,255}
\definecolor{companiondark}{RGB}{220,235,255}
\definecolor{flightlight}{RGB}{240,255,240}
\definecolor{flightdark}{RGB}{220,255,220}
\definecolor{hardwarelight}{RGB}{255,250,240}
\definecolor{hardwaredark}{RGB}{255,240,220}
\definecolor{actuatorlight}{RGB}{255,240,240}
\definecolor{actuatordark}{RGB}{255,220,220}
\definecolor{spraylight}{RGB}{250,240,255}
\definecolor{spraydark}{RGB}{235,225,255}

\IEEEoverridecommandlockouts
\usepackage{hyperref}
\hypersetup{urlcolor=DBlue, 
citecolor=DBlue, 
linkcolor=DBlue, 
colorlinks=true}

\title{Reducing the Power Required for Multi-UAV Tethered Payload Transportation via a Rotating Equilibrium}

\author{Eric Foss$^*$, Andrew Tai$^*$, Carlo Bosio$^*$, Mark W. Mueller%
\thanks{$^*$Equal contribution.}
\thanks{The authors are with High Performance Robotics Laboratory, Department of Mechanical Engineering, University of California Berkeley, CA 94720, United States.}
}

\begin{document}
\maketitle

\begin{abstract}

Collaborative aerial transportation of tethered payloads is fundamentally limited by space, power, and weight constraints. 
Conventional approaches rely on static equilibrium conditions, where each vehicle tilts to generate the forces that ensure they maintain a formation geometry that avoids aerodynamic interactions and collision. 
This horizontal thrust component represents a significant energy penalty compared to the ideal case in which each vehicle produces purely vertical thrust to lift the payload. 
Operating in tighter tether configurations can minimize this effect, but at the cost of either having to fly the vehicles in closer proximity, which risks collision and aerodynamic interactions, or significantly increasing the length of the tether, which increases complexity and reduces potential use-cases. 
We propose operating the tether-suspended flying system at a rotating equilibrium. 
By maintaining steady circular motion, centrifugal forces provide the necessary horizontal tether tension, allowing each quadcopter to generate purely vertical thrust and thus reducing the total force (and power) required compared to an equilibrium where the thrusts are not vertical. 
It also allows for a wider range of tether configurations to be used without sacrificing efficiency.
Results demonstrate that the rotating configuration can reduce power consumption relative to static lifting by up to \changed{$18.4\%$}, making collaborative aerial solutions more practically relevant. 
\changed{The system is also demonstrated in outdoor tests, showcasing control robustness in uncertain environments. }

\end{abstract}



\section{Introduction}
Collaborative aerial transport enables teams of multicopters to carry payloads exceeding the capability of individual vehicles. 
Such systems are attractive for logistics, construction, inspection, and emergency response applications \cite{villa2020survey}. 
The two most prominent approaches to collaborative aerial transport mainly differ in the way vehicles are attached to the payload, i.e. tethered connections \cite{estevez2024review, zhang2023formation, sun2025agile} or rigid attachments \cite{mellinger2013cooperative, oishi2021autonomous, bosio2025automated}. The tethered strategy allows the vehicles to distance from each other and the payload, avoiding undesired aerodynamics interactions, but it is subject to complex and often unwanted internal swing dynamics. The rigid connection strategy allows more precise control, but is typically more sensitive to hard-to-model payload vibrational modes \cite{ritz2013carrying}. 
\begin{figure}[tb]
    \centering
    \includegraphics[width=0.99\linewidth]{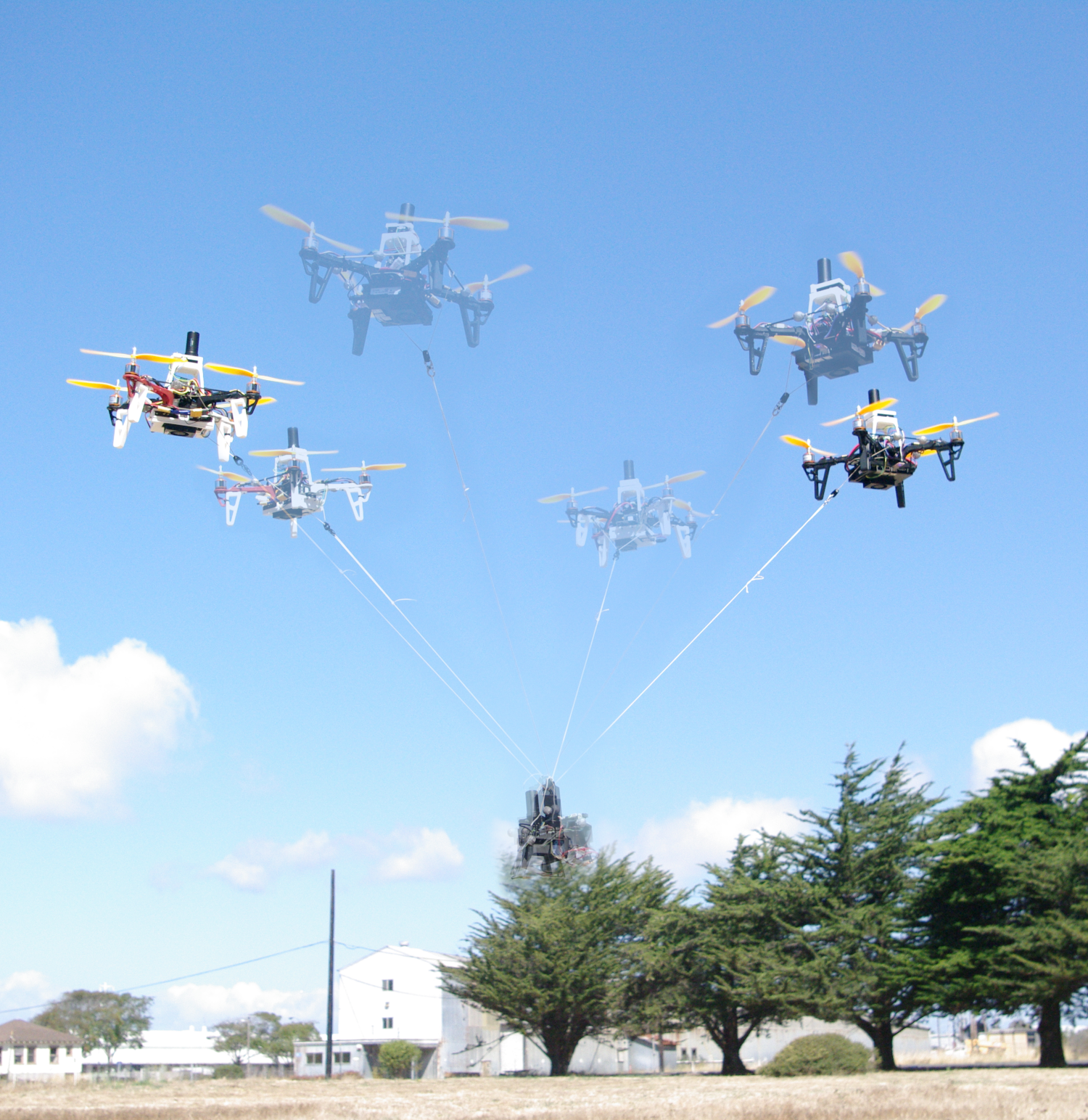}
    \caption{Composite image showing a rotating payload transport system \changed{during an outdoor experiment.} As it can be observed, even with large tether angles, the vehicle thrust axes are nearly vertical. In this particular experiment, the tether angle $\beta=45^\circ$ from vertical, and the vehicles are rotating with a tangential velocity of \changed{$\sim1.9\,\mathrm{m\cdot s^{-1}}$}.}
    \label{fig:overlay}
\end{figure}

However, the flight endurance achieved by all these approaches remains severely limited by battery capacity, making energy efficiency a central design consideration. 
In fact, flight endurance is one of the main issues limiting the deployment of aerial robots for transportation. Several ideas have been proposed to increase flight endurance of multicopters \cite{alyassi2022autonomous}. Example methods of practical use include power beaming \cite{lahmeri2022charging} and in-air battery recharging \cite{jain2020flying}. 

Our main observation is that, in most previous work on tether-suspended payload transport, relative motion between the payload and the vehicles is mostly avoided unless necessary for tasks such as obstacle avoidance \cite{jackson2020scalable}. We denote this standard approach as ``static''. In static tether-suspended cooperative transport, each multicopter must generate both vertical lift to suspend the payload and additional horizontal force to maintain formation geometry.
While mechanically intuitive, this strategy consumes more power compared to an energetically optimal situation in which the vehicles generate purely vertical thrusts. This loss in power efficiency can be reduced by operating in tight tether configurations; however, the reduction in the distance between the carriers leads to higher risks of collision. This reduction in vehicle distance can be mitigated with a longer tether, but this comes at the cost of increased mass, more complex tether dynamics, and increased size, making these systems less practical. Thus, we propose a multicopter tether-suspended transport control method that eliminates the need for this additional horizontal force through the use of a rotating equilibrium configuration, which can also be considered as a relaxed hover condition \cite{mueller2016relaxed} for a tether-suspended system.

The use of rotating equilibria to vertically lift payloads has been studied in the case of fixed-wing aircraft, motivated by the fact that they require high airspeeds to maintain stable flight, unlike traditional quadcopters. By orbiting at high speeds above the payload, fixed-wing aircraft take advantage of their aerodynamic surfaces to generate high lift forces, providing a weight-efficient method to transport heavy payloads. Early work featured dynamic analysis of large-scale aircraft necessitating extremely long cable lengths, increasing the need for complex aerodynamic tether modeling \cite{WilliamsTether, Williams2007DynamicsPart1, Williams2012Multi}. Later works reduced the scale of the system, utilizing advancements in small-scale fixed-wing hardware and control methods to create an experimental sub-scale version \cite{IndoorExperiment, Quenneville2023Experimental}. Previous work has also studied high-speed, tether-constrained quadcopter flight, but not for payload transportation \cite{schulz2015high}. 

Our work builds upon these ideas by expanding them to a collaborative tether-suspended multicopter system \changed{with a single attachment point.} 
By maintaining steady circular motion around a vertical axis, the centrifugal acceleration can provide the necessary horizontal components of the tether tension required to maintain a desired tether configuration that keeps a safe distance between the vehicles. 
In this configuration, each multicopter can then produce purely vertical thrust equal to its own weight plus its payload weight share (see Fig. \ref{fig:overlay} for an example), reducing its steady-state power consumption. 
\changed{While this is also potentially achievable through the use of multiple attachment points on a large payload, this increases mechanical and modeling complexity, and the single attachment point method could be more generalizable to a wide array of potential payloads.}
It also unlocks the ability to operate in flatter tether configurations without requiring significantly more power, allowing for shorter tether lengths without sacrificing close-proximity safety concerns.
The cost of this strategy is additional aerodynamic drag due to forward motion, however, it has been shown that forward flight can be more efficient than hovering when flying at low speeds \cite{helicopteraero}, \cite{levelflightpower}, \cite{swarmpoweranalysis}. 
We show that for a range operating regimes, the drag-induced power remains small compared to the thrust penalty incurred under static equilibrium.

The main contributions of this work are:

\begin{itemize}
    \item A first-principles comparison of static and rotating equilibria for tether-suspended payload transport.
    \item A control strategy for stabilizing rotating equilibria.
    \item Experimental validation demonstrating energy savings in experimental flight tests.
\end{itemize}




\section{Modeling and control}
\begin{figure}
    \centering
    \includegraphics[width=0.99\linewidth]{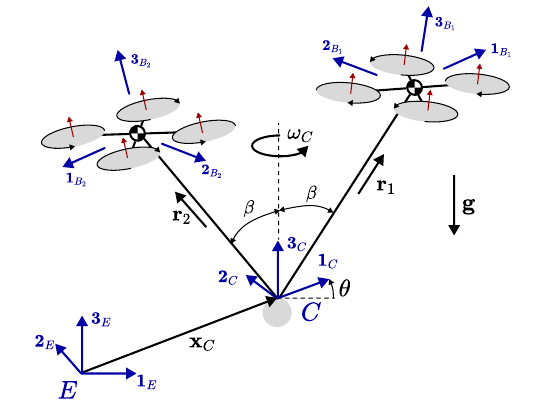}
    \caption{Diagram of multi-UAV tethered system transporting a point-mass payload with $N_q = 2$. \changed{The $C$ reference frame origin is defined as the desired payload position}, and the two quadcopters lie on the plane defined by $\text{span}(\mathbf{1}_C, \mathbf{3}_C)$. The $C$ frame rotates with respect to the earth-fixed frame $E$ with angular velocity $\omega_C \mathbf{3}_C$, and its orientation with respect to $E$ is described by a rotation by $\theta$ about $\mathbf{3}_E$. When $\omega_C = 0$ the system is a classical, static dual-UAV transport system.}
    \label{fig:sys-drawing}
\end{figure}
We consider \changed{$N_q \geq 2$} identical \changed{point-mass} quadcopters of mass $m_q$ transporting a point-mass payload of mass $m_p$ via massless tethers of length $\ell$. \changed{For the following analysis, we assume the quadcopters are equally spaced around the payload, resulting in the quadcopters sharing the weight of the payload equally and therefore optimally.} We introduce an earth-fixed \changed{East-North-Up} coordinate frame $E$ which we assume to be inertial. With reference to Fig.~\ref{fig:sys-drawing}, we also introduce a control frame $C$ with the position of its origin denoted as $\mathbf{x}_C$, which coincides with the desired position of the payload. The control frame rotates at a constant rate $\boldsymbol{\omega}_C$ = $[0, 0, \omega_C]^T$ about a shared vertical axis with the inertial frame. \changed{A body-fixed frame $B_i$ is also defined for each quadcopter, centered at its center of mass.} We denote $\mathbf{x}_{i}$ as the position of quadcopter $i$, $\mathbf{x}_p$ as the position of the payload. $\hat{\mathbf{r}}_{i}$ represents the unit vector pointing from the payload to quadcopter $i$, so that $\mathbf{r}_{i} = \ell \hat{\mathbf{r}}_{i}$ is the position of quadcopter $i$ with respect to the payload. \changed{The tether angle $\beta$ is defined as the angle between $r_i$ and the vertical axis $\mathbf{3}_C$, and is assumed to be nominally equal for each quadcopter.}

We denote with a superscript the components of a vector in a given reference frame. For example $\mathbf{v}^E$ describes the components of the vector $\mathbf{v}$ in the reference frame $E$. \changed{The point-mass modeling of the quadcopters assumes the tethers are attached at the center of mass}, simplifying the analysis by decoupling the vehicles' linear dynamics from their attitude dynamics. This configuration allows position and attitude controllers to be separated in a cascaded architecture.

\subsection{System Dynamics}
We write the quadcopters' linear dynamics in the Earth-fixed frame as

\begin{equation} \label{eq:quad_dyn}
    m_q \ddot{\mathbf{x}}_i^E = T_i \changed{\mathbf{e}_{3_{B_i}}^E} + m_q \mathbf{g}^E + \mathbf{f}_{d, i}^E - F_i \hat{\mathbf{r}}_i^E,
\end{equation}

where $i=1, \ldots, \changed{N_q}$ is the vehicle index, $T_i$ is the total thrust produced by vehicle $i$, \changed{$\mathbf{e}_{3_{B_i}}^E$ is a unit vector in body three direction expressed in reference frame $E$}, $F_i$ is the tether tension between the payload and vehicle $i$, $\mathbf{g}$ is the acceleration due to gravity ($\mathbf{g}^E = [0,0,-9.81]^T$), and $\mathbf{f}_{d,i}$ is the drag force experienced by vehicle $i$. 

The payload's linear dynamics are in the form
\begin{equation} \label{eq:payload_dyn}
    m_p \ddot{\mathbf{x}}_p^E = m_p\mathbf{g}^E + \mathbf{f}_{d, p}^E + \sum_i F_i \hat{\mathbf{r}}_i^E,
\end{equation}
where $\mathbf{f}_{d,p}$ is the drag force experienced by the payload.

For simplicity in the dynamics formulation, we model the tethers as very high stiffness springs with \changed{rest length $\hat{\ell}$}, stiffness $k_T$, and damping $c_T$. The tether force characterization which provides a dynamic link between the vehicles and payload is

\begin{equation} \label{eq:tether_dyn}
F_i = k_{T} \left( \lVert \mathbf{r}_{i} \rVert - \changed{\hat{\ell}} \right) + c_{T} \left( \frac{d}{dt}{\lVert \mathbf{r}_{i} \rVert } \right).
\end{equation}

\subsection{Equilibrium Conditions}

We are interested in equilibria relative to a rotating control frame $C$ assuming the quadcopters are equally spaced about the payload, resulting in the following equilibrium conditions:

\begin{align} \label{eq:eq_cond}
    \begin{cases}
    \bar{\dot{\mathbf{x}}}_C^E = \text{constant} \\
    \bar{\ddot{\mathbf{x}}}_C^E = 0 \\
    \bar{\dot{\mathbf{r}}}_{i}^C = 0 \\
    \bar{\ddot{\mathbf{r}}}_{i}^C = 0 ,
    \end{cases}
\end{align}
where $\mathbf{x}_C^E$ represents the position of the control frame $C$ relative to the Earth-fixed frame, and $\mathbf{r}_{i}^C$ represents the position of quadcopter $i$ with respect to the the payload position $\mathbf{x}_p$, which in equilibrium is at the origin of the control frame. 

Note that these equilibrium conditions do not result in a static system with respect to the inertial frame. For brevity, we show this by analyzing quadcopter $1$ in the system shown in Fig. \ref{fig:sys-drawing}. The position of quadcopter $1$ in the inertial frame is
\begin{equation} \label{eq:quad1_pos}
    \mathbf{x}_{1}^{E} = \mathbf{x}_{C}^{E} + R_{C}^E \mathbf{r}_{1}^{C}, \quad
    \mathbf{r}_1^C = \ell \begin{bmatrix}
        \sin \beta \\
        0          \\
        \cos \beta
    \end{bmatrix},
\end{equation}
where $R_C^E$ represents the coordinate rotation from control frame $C$ to earth-fixed frame $E$, \changed{and $\ell$ is the equilibrium stretched length of the spring, which due to its high stiffness is assumed to stay the length of the tether.} Taking the first time derivative of \eqref{eq:quad1_pos} yields:
\begin{equation}
    \dot{\mathbf{x}}_{1}^{E} = 
    \dot{\mathbf{x}}_{C}^{E} + R_{C}^{E} \dot{\mathbf{r}}_{1}^{C} + \dot{R}_{C}^{E} \mathbf{r}_{1}^{C}.
\end{equation}
Noting the second equilibrium conditions of \eqref{eq:eq_cond}, we can see the equilibrium velocity of quadcopter $1$ with respect to the moving control frame $C$, represented in the control frame coordinate system, is 
\begin{equation} \label{eq:quad1_vel}
\bar{\dot{\mathbf{x}}}_{1}^{E} - \bar{\dot{\mathbf{x}}}_{C}^{E} = 
\boldsymbol{\omega}_{c} \times \mathbf{r}_{1}^{C} = 
\begin{bmatrix}
0 \\
\omega_{c} \ell \sin \beta \\
0
\end{bmatrix} .
\end{equation}
Taking the second time derivative of \eqref{eq:quad1_pos} (omitted for brevity) and noting the first and third equilibrium conditions of \eqref{eq:eq_cond}, we derive the equilibrium acceleration of quadcopter $1$ with respect to the moving control frame $C$, represented in the control frame coordinate system, is 

\begin{equation} \label{eq:quad1_acc}
\bar{\ddot{\mathbf{x}}}_{1}^{E} = \begin{bmatrix}
-\omega_{C}^{2} \ell \sin \beta \\
0 \\
0
\end{bmatrix}.
\end{equation}
Note for quadcopter $2$, the equilibrium velocity and acceleration with respect to the moving control frame $C$ is the negative of \eqref{eq:quad1_vel} and \eqref{eq:quad1_acc} respectively, and can be obtained by setting $\mathbf{r}_2^C = \ell [
        -\sin \beta,
        0,
        \cos \beta]^T$. 

Substituting the first equilibrium condition of \eqref{eq:eq_cond} into the dynamics of the payload shown in \eqref{eq:payload_dyn} yields:
\begin{equation}
    \mathbf{0} = m_p \mathbf{g} + \mathbf{f}_{d, p} + \sum_i F_i\mathbf{r}_i.
\end{equation}
We assume the quadcopters are equally spaced above the payload and the payload's drag is minimal due to low translational velocities, and solve for the tether tension at equilibrium:
\begin{equation}
    \bar{F}_1 = \changed{ \cdots = \bar{F}_{N_q}} = \frac{\changed{m_p} g}{\changed{N_q} \cos \beta}.\label{eq:eq-forces}
\end{equation}

\subsection{Control\label{sec:control}}
\begin{figure}
    \vspace{0.2cm}
    \centering
    \includegraphics[width=0.99\linewidth]{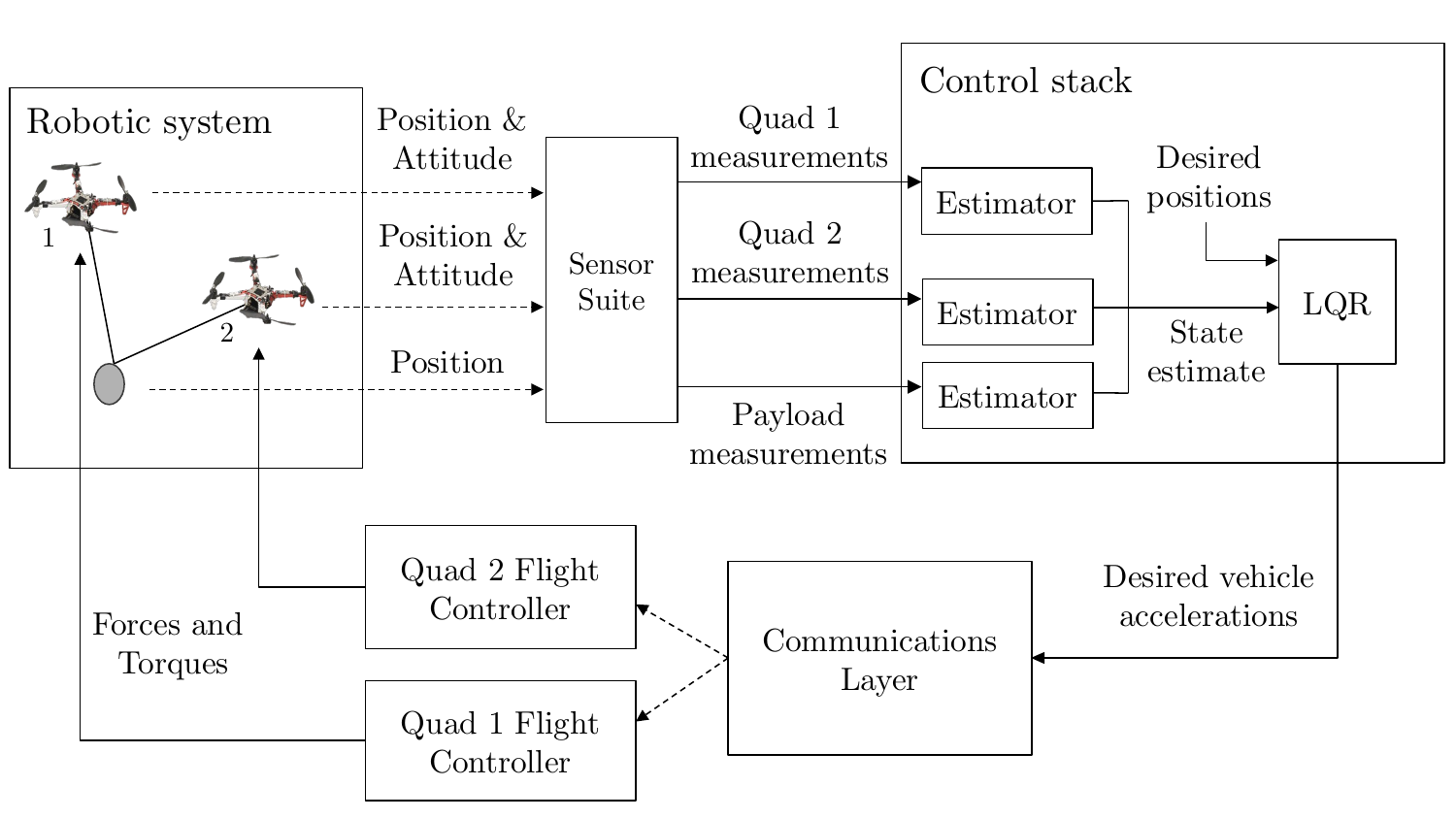}
    \caption{Block diagram of the system architecture. \changed{The sensor suite is a motion capture system (indoors) and a RTK-enabled GPS unit (outdoors), running at $200\,\mathrm{Hz}$ and $10\,\mathrm{Hz}$ respectively. The estimators run at $200\,\mathrm{Hz}$. The LQR controller runs at $50\,\mathrm{Hz}$, while the quadcopters' onboard flight controllers run at $500\,\mathrm{Hz}$, providing an effective separation of timescales.}}
    \label{fig:blk-diag}
\end{figure}
The previous dynamics derivation allows us to linearize the system about equilibrium conditions and design a controller for the linearized system. In particular, assuming that the quadcopters' attitude dynamics is characterized by much lower time constants than the three-mass system linear dynamics, we can design a high level position controller to control the thrust vectors of the vehicles.

For this, we design a Linear Quadratic Regulator (LQR) on the first-order system model expressed in the (non-inertial) $C$ frame. The LQR controller outputs a desired acceleration for each vehicle. We assume that the vehicle low-level control can achieve a target acceleration and thrust value nearly instantaneously compared to the time constant of the outer loops. The state $\mathbf{s}\in\mathbb{R}^{\changed{6(N_q+1)}}$ consists of the positions and velocities of the payload and quadcopters relative to $E$ expressed in components in the control frame, i.e. 
\begin{equation}
\mathbf{s} = \left[
        \mathbf{x}_{p}^C,
        \dot{\mathbf{x}}_{p}^C,
        \mathbf{x}_{1}^C,
        \dot{\mathbf{x}}_{1}^C,
        \changed{
        \ldots,
        \mathbf{x}_{N_q}^C,
        \dot{\mathbf{x}}_{N_q}^C}\right]^T.
\end{equation}        
The control input vector $\mathbf{u}\in \mathbb{R}^{\changed{3N_q}}$ contains \changed{$N_q$} desired thrust vectors, one per quadcopter, i.e.
\begin{equation}
    \mathbf{u} = \left[
        \mathbf{T_1},
        \changed{
        \ldots,
        \mathbf{T_{N_q}}}\right]^T,
\end{equation}
where $\mathbf{T}_i\in\mathbb{R}^3$ represents a thrust vector generated by vehicle $i$ represented in the control frame coordinate system.
The LQR feedback components are combined with the feedforward terms given by the linearization to give a desired acceleration command. This is then tracked by a lower-level controller. 

The LQR feedback matrix $K$ directly depends on the system's physical parameters ($m_q$, $m_p$, $\ell$), the desired flight conditions ($\omega_C$, $\beta$), and the LQR cost 
\begin{equation}
J = \int_{t=0}^{\infty}(\mathbf{s}^TQ\mathbf{s} + \mathbf{u}^TR\mathbf{u})dt,
\label{eq:lqr}
\end{equation}
where
\begin{equation}
    Q \in \mathbb{R}^{\changed{6(N_q+1) \times 6(N_q+1)}}, \quad R \in \mathbb{R}^{\changed{3N_q \times 3N_q}}.
    \label{eq:lqrcosts}
\end{equation}

A mid-level attitude control then compares the direction of this generated desired acceleration vector with the attitude estimate and generates a desired angular velocity vector. \changed{An integrator is also used at this level to correct for unmodeled moment from the tether due to potential error between the tether attachment point and center of mass.} A low-level angular-velocity controller then compares this with the onboard inertial measurement unit readings to generate motor commands. We design the inner loops to be faster than the outer loops to achieve a wide enough separation of time scales. A block diagram describing the system logic is shown in Fig. \ref{fig:blk-diag}. 


\section{Power analysis\label{sec:pow-analysis}}
In this section we provide a first-principles comparison of the power consumption of a rotating system compared to a static system.
\begin{figure}[tb]
  \begin{center}
    \includegraphics[width=0.25\textwidth]{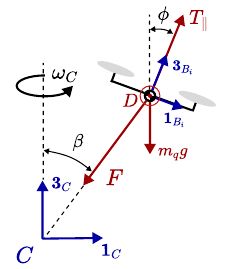}
  \end{center}
  \caption{Free body diagram of a quadcopter in the rotating frame. \changed{The drag force $D$ is a symptom of a constant tangential velocity due to the rotating equilibrium, and thus points out of the page at the center of mass of the quadcopter. The thrust vector depicted $T_{\parallel}$ is the component of the total thrust within the plane. The total thrust vector also has a component $T_{\perp}$ pointing into the page, opposing the drag vector.}}
  \label{fig:planar-eq}
\end{figure}
We analyze more in depth the system equilibrium conditions~\eqref{eq:eq_cond} in the $\mathbf{1}_C-\mathbf{3}_C$ plane for a single quadcopter, as depicted in Figure \ref{fig:planar-eq}. $T_{\parallel}$ represents the component of quadcopters total thrust vector in the plane while $\phi$ represents the angle of this vector from the vertical $\mathbf{3}_C$ direction, and $F$ represents the tether tension. 

\subsection{Equilibrium Force Balance}

The equilibrium conditions in the horizontal, vertical, \changed{and out-of-plane directions} for the quadcopter, as well as the vertical direction for the payload, can be written as
\begin{align}
    \begin{cases}
            T_{\parallel}\cos \phi - F \cos \beta - m_q g = 0\\
            T_{\parallel} \sin \phi - F \sin \beta = -m_q \omega_C^2 \ell \sin \beta \\
            \changed{T_{\perp} - D = 0} \\
            \changed{N_q} F \cos \beta - m_p g = 0,
    \end{cases}
\end{align}
leading to the equations:
\begin{align}
    \begin{cases}
            F = \frac{m_p g}{\changed{N_q} \cos \beta} \\
            T_{\parallel}\cos \phi = F\cos \beta + m_q g\\
            T_{\parallel} \sin \phi = F\sin \beta -m_q \omega_C^2 \ell \sin \beta \\
            \changed{T = \sqrt{T_{\parallel}^2 + D^2}.}
    \end{cases}
    \label{eq:static-eq}
\end{align}
The second and third equation in \eqref{eq:static-eq} quantify respectively the vertical and horizontal component of the thrust required to keep the system in equilibrium. \changed{The fourth equation combines in-plane thrust $T_{\parallel}$ with the out-of-plane thrust $T_{\perp}$ opposing the drag force to obtain the total thrust $T$.} The magnitude of $T_{\parallel}$ is given by

\begin{align}
    T_{\parallel} &= \sqrt{\left(\frac{m_p g}{\changed{N_q}} + m_q g\right)^2 + \left(\sin \beta \left(\frac{m_p g}{\changed{N_q} \cos \beta} - m_q \omega_C ^2 \ell \right) \right)^2}.\label{eq:T_general}
\end{align}

It is interesting to note that the vertical component does not depend on $\beta$ or $\omega_C$, rather only the total system weight. The horizontal component, however, depends on both. Trivially, we note that the horizontal component is $0$ when $\beta = 0$ (i.e., when the quadcopters would collide with each other). Non-trivially, for any non-zero value for $\beta$, there exists a critical angular velocity value $\omega^*_C$ which results in zero in-plane horizontal thrust:
\begin{equation}
    \omega_C^* = \sqrt{\frac{m_p g}{\changed{N_q} m_q \ell \cos \beta}}.
    \label{eq:optimal-omega}
\end{equation}

Operating the system at the critical angular velocity results in a purely vertical in-plane thrust of magnitude 
\begin{equation}
    T_{\parallel}^* = \frac{m_p g}{\changed{N_q}} + m_q g.
\end{equation}
On the other hand, operating the system in the static configuration ($\omega_C = 0$) results in a non-vertical in-plane thrust of magnitude
\begin{equation}
    T_{\parallel}^s= \sqrt{\left(\frac{m_p g}{\changed{N_q}} + m_q g\right)^2 + \left(\frac{m_p g}{\changed{N_q}}\tan \beta\right)^2}.\label{eq:T-static}
\end{equation}

\changed{The drag force is modeled as}
\begin{equation}
    \changed{D = \frac{1}{2} \rho v_{\perp}^2 S C_D, \quad v_{\perp} = \omega_C \ell \sin \beta} 
\end{equation}
\changed{where $v_{\perp}$ is the tangential, out-of-plane velocity. We note that for the static case $D = 0$, resulting in $T^s = T_{\parallel}^s$.}

%


\subsection{Power Consumption}

To analyze the effect this has on power consumption, we use actuator disk theory \cite{helicopteraero} to compare the required power to hold each equilibrium configuration. 


For near hover conditions, the following relation can be used to estimate the power consumption to maintain hover:

\begin{equation}
    P_{h,\text{ideal}} = \frac{T_h^{3/2}}{r_p\sqrt{2 \pi \rho N_p}}, \label{eq:power-consmp}
\end{equation}
where $P_{h,\text{ideal}}$ is the ideal power at hover, $T_h$ is the total thrust required at hover, $\rho$ is the air density, $r_p$ is the rotor radius, and $N_p$ is the number of propellers.

\changed{However, in rotating conditions the quadcopters are traversing in level flight, resulting in a non-axial component of airflow through the propellers. In this case, we use Glauert's Momentum Theory for Forward Flight \cite{helicopteraero} to estimate the induced velocity as}
\begin{equation}
    \changed{
    v_i = \frac{T}{ 2 \rho \pi r_p^2 N_p\sqrt{(v_{\perp} \cos \alpha)^2 + (v_{\perp} \sin \alpha + v_i)^2}},
    }
    \label{eq:impicit-induced-vel}
\end{equation}
\changed{where $v_i$ is the induced velocity, and $\alpha$ is the angle between the rotor and level plane. For both the static and rotating case, we assume the thrust is shared equally between all propellers. Note that the induced velocity $v_i$ appears on both sides of the equation, and must be solved for using a root-solving numerical method such as Newton-Raphson. In level flight, the angle $\alpha$ represents the angle the quadcopter must tilt forward to oppose drag, and is given by}
\begin{equation}
    \changed{\alpha = \tan^{-1}\left(\frac{D}{m_q g} \right).}
\end{equation}
\changed{The induced power consumption for level flight is then given by}
\begin{equation}
    \changed{P_{l, \text{ideal}} = Tv_i.}
\end{equation}

\changed{The above power estimates for hover and level flight are for ideal power consumption, and therefore do not consider the profile drag and other viscous effects of the propellers, the power to overcome the parasite drag, and the powertrain efficiency. The ideal power estimates are thus first corrected using the propeller's Figure of Merit $FoM$ to accurately estimate the true shaft power of the rotors, correcting for the unmodeled profile drag and other non-ideal effects: }
\begin{equation}
    \changed{P_{\text{shaft}} = \frac{P_{\text{ideal}}}{FoM}.}
\end{equation}
\changed{Then, the power of the parasitic drag is added to the shaft power before corrected once again by the powertrain efficiency $\eta_{\text{powertrain}}$:}
\begin{equation}
    \changed{P_{\text{est}} = \frac{P_{\text{shaft}} + P_{\text{drag}}}{\eta_{\text{powertrain}}}, \quad P_{\text{drag}} = D v_{\perp}}
\end{equation}



\begin{figure}
    \centering
    \includegraphics[width=0.8\linewidth]{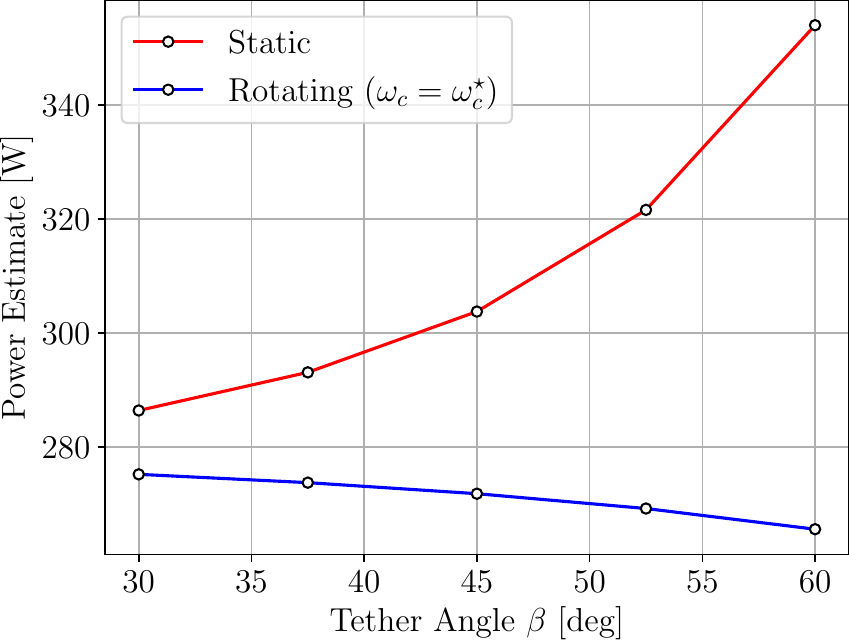}
    \caption{\changed{Plot of predicted power consumption of a single quadcopter in the experimental system (Tab. \ref{tab:exp-params}) versus tether angle for both the static case and the rotating case, where $\omega_C = \omega_C^*$.}}
    \label{fig:power-beta-plot}
\end{figure}

\begin{figure}
    \centering
    \includegraphics[width=0.8\linewidth]{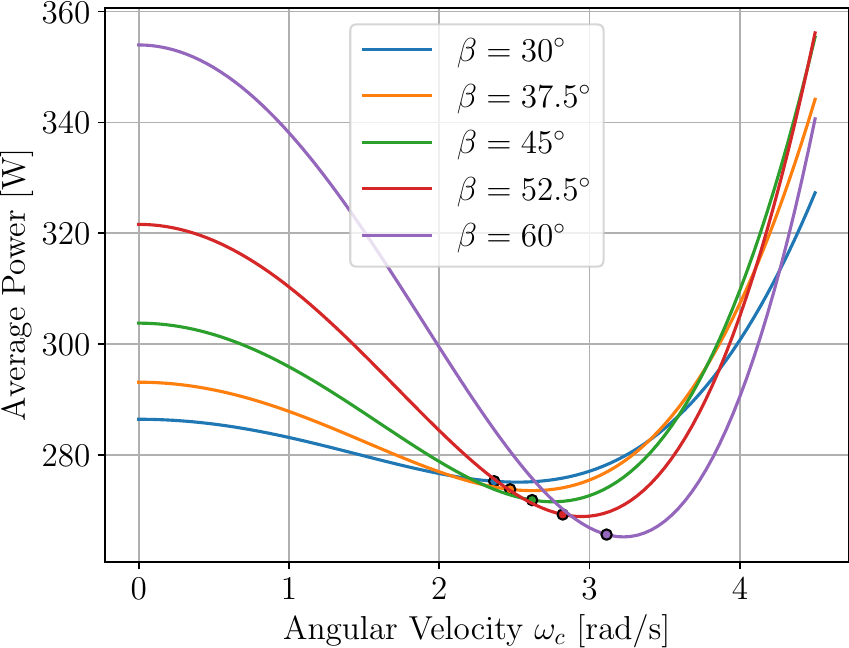}
    \caption{\changed{Plot of predicted power consumption of a single quadcopter in the experimental system (Tab. \ref{tab:exp-params}) as a function of the rotating frame angular velocity. The point on each line represents when $\omega_C = \omega_C^*$, given by Equation \eqref{eq:optimal-omega}.}}
    \label{fig:power-omega-plot}
\end{figure}









\subsection{Analytical findings}

Using the experimental parameters, we can calculate the power required for hover for both the rotating and static systems at a specified $\beta$. For the rotating system, we set $\omega_C = \omega_C^*$. 

\changed{Fig.~\ref{fig:power-beta-plot} demonstrates that for the rotating configuration, power consumption decreases as the tether angle increases and thus tangential velocity $v_{\perp}$. This corresponds with the well-documented behavior that low-speed level flight can require less power than hover up to certain speeds \cite{helicopteraero}, \cite{levelflightpower}, \cite{swarmpoweranalysis}. For our system parameters this effect is small, but for systems with heavier payloads this effect could potentially result in further power reduction. There also potentially exists system configurations where the tangential velocity exceeds the low-speed power saving region, resulting in an increasing rather than decreasing trend. Furthermore, some experimental results on quadcopters show a flat, rather than decreasing, trend for low-speeds \cite{experimentallevelflightquadcopters}.} In contrast, the power consumption for the static case increases monotonically with increasing $\beta$. 

\changed{Fig.~\ref{fig:power-omega-plot} further demonstrates that for any $\beta > 0$, there exists a minimum near $\omega_C = \omega_C^*$ that is the lowest power consumption configuration for the system. This discrepancy is due to the fact that $\omega_C^*$ is the value that minimizes the in-plane thrust component $T_{\parallel}$, not the total thrust $T$ which depends also on the parasite drag $D$. The optimal $\omega_C$ that minimizes the average power appears to be slightly greater than $\omega_C^*$ due to the forward-flight power saving behavior mentioned previously, which makes it difficult to find due to the implicit nature of Equation \eqref{eq:impicit-induced-vel}.} If $\omega_C^*$ is greatly exceeded, then the vehicle must generate inward pointing thrust to produce the required centripetal acceleration, thus increasing power consumption.

Further motivation for this architecture can be observed when payload mass is large relative to vehicle mass. As shown in the first expression in \eqref{eq:static-eq}, the horizontal component of tether force required to keep system equilibrium in the static case scales with increasing payload mass, highlighting the advantage of the using the rotating equilibrium. 

\section{Experiments\label{sec:exp}}

We validate our power consumption predictions through experimental flight tests. The experiments consisted of equilibrium hover flights, in both static and rotating conditions. \changed{To minimize the effects of external disturbances like wind, or significant sensor uncertainty, we report the power consumption data using flights in an indoor motion capture space. However, we also showcase robustness and tracking in uncertain environments with an outdoor flight demonstration.}
\begin{table}[tb]
\normalsize
    \centering
    \caption{\footnotesize{Parameters of the system used in experiments.}}
    \begin{tabular}{c|c}
        \changed{Quad. mass }$m_q$ & $1.06\,\mathrm{kg}$ \\
        Number of quads $N_q$ & $2$ \\ 
        Number of propellers $N_p$ & $4$ \\ 
        Propeller radius $r_p$ & $0.1\,\mathrm{m}$ \\ 
        \changed{Payload mass} $m_p$ & $1.18\,\mathrm{kg}$ \\
        Tether length $\ell$ & $1\,\mathrm{m}$ \\
        \changed{Figure of Merit} $FoM$ & $0.575$ \\
        \changed{Powertrain Efficiency} $\eta_{\text{powertrain}}$ & $0.725$ \\
        \changed{Parasite Drag Term} $S \times C_D$ & $0.047\,\mathrm{m^2}$
        
    \end{tabular}
    \label{tab:exp-params}
\end{table}

\subsection{System Design}

\begin{figure}
    \centering
    \includegraphics[width=0.8\linewidth]{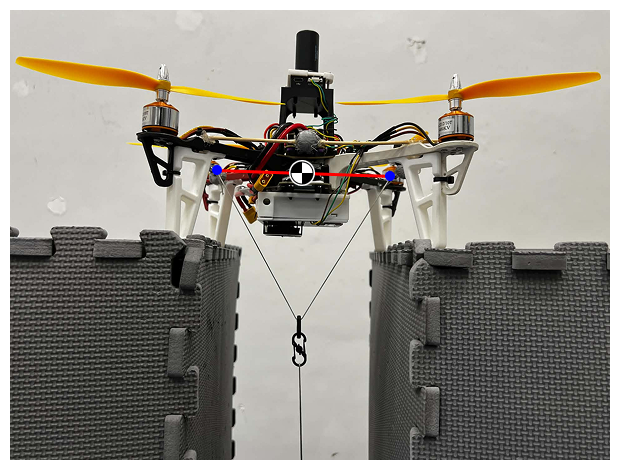}
    \caption{\changed{Experimental hardware setup showing tether attachment passing through center of mass.}}
    \label{fig:exp-hardware-front}
\end{figure}

Physically attaching the tethers at the center of mass of vehicles and payload, a key assumption in the dynamics model, is practically very challenging. For the payload, we found that this is not strictly required, and attaching the tethers above the center of mass reduces rotational payload oscillations. For the quadcopters, this condition is more critical, as any offset between the force vector and the vehicles' center of mass would generate a large moment. \changed{To match the model assumption as close as possible, two rods were mounted on the quadcopter arms, parallel to $\mathbf{1}_{B_i}$ (see Fig. \ref{fig:sys-drawing}). These rods are on the same $z$ height as the quadcopter center of mass, as shown in Fig. \ref{fig:exp-hardware-front}. A string connects to the midpoints of the two rods, and is attached to the rest of the length of tether through a slider. During flight, the tethers are angled such that they pivot about the $\mathbf{2}_{B_i}$ axis, leading to the tether force passing through the quadcopters' center of mass exactly. However, this requires the quadcopters to yaw with the rotating frame. The slider is an ellipse, so for rotations about the $\mathbf{1}_{B_i}$ direction (i.e. tilting forwards tangent to the circle), the tether force no longer acts along the center of mass. This offset, however, is small.}



\changed{The propeller Figure of Merit and the quadcopter's powertrain efficiency were found experimentally using a \textit{Tyto Robotics Series 1580 Thrust Stand}. For a full range of thrust setpoints, the electrical power, shaft power, and ideal power were compared. The electrical power was measured directly via the test stand's current and voltage sensors. The shaft power was also measured directly via motor torque and optical speed measurements. Equation \eqref{eq:power-consmp} was used alongside the thrust measurements to estimate the ideal power of the propeller. The ideal and shaft power were compared to estimate the Figure of Merit, and the shaft and electrical power were compared to estimate the powertrain efficiency. The parasite drag term was estimated using a flat plate model with the cross-sectional area of the quadcopter frame and electronic peripherals. The resulting values can be found in Table \ref{tab:exp-params}.}

\changed{A \textit{B-CUBE Power Module PRO} was used to measure the power consumption of the quadcopter. It was calibrated by powering the quadcopter from a bench-top power supply and running the motors at 50 \% the quadcopter's weight. The sensor's current and voltage readings are then compared against the ground-truth power supply values to fit a linear corrective term.  } 
Additionally, we subtracted the steady state power (the power consumed by the flight controller and ESCs while the motors are stationary), as that also is not included in the analytical model.

\changed{The LQR costs (see Equation \eqref{eq:lqr})  were first tested in simulation, then further tuned when running initial indoor and outdoor experiments. We achieved good indoor performance with the quadcopter and payload $x,y,z$ position and velocity error costs in the $Q$ matrix set to $15, 7.5, 15, \text{and} \, 0$ respectively. Outdoors, the costs had to be reduced to account for the less optimal estimation to $8, 4, 8, \text{and} \, 0$ respectively. The $x,y,z$ control costs for each quadcopter in the $R$ matrix were set to $1.2, 1.2, \text{and} \, 1.0$ respectively. }

\changed{For outdoor estimation, both the quadcopters and payload use \textit{Holybro H-RTK F9P Helical} GPS modules, which receive RTK corrections from a static base station. Each quadcopter uses a Pixracer R15 flight controller that runs an Extended Kalman Filter (EKF) to fuse 500 Hz IMU readings, and 10 Hz GPS or 200 Hz motion capture data, for outdoor and indoor experiments respectively. Additionally, onboard \textit{Jetson Orin Nanos} serve as the communication layer—broadcasting state estimates and telemetry while routing RTK corrections, outer-loop control, and user commands to the flight controller at 100 Hz. The mission state machine and LQR outer-loop controller runs offboard on a computer at 50 Hz. A diagram of the described system can be found in Fig. 3.}

\subsection{Experimental Results}

Experimental data was obtained via equilibrium hover tests for both the static and rotating configuration in an indoor motion capture space. The experiments consisted of take-off, static hover for $10\,\mathrm{s}$, spin-up to $\omega_C^*$, then hold in the rotational equilibrium for $20\,\mathrm{s}$. Five tests were run at different tether angles $\beta$, spanning from $30^\circ$ to $60^\circ$. The power measurement is averaged over the $10\,\mathrm{s}$ static period and $20\,\mathrm{s}$ rotational period respectively. Videos of the experiments can be found in the supplementary video.

Fig~\ref{fig:exp-results} shows the recorded power consumption between the static and rotating equilibrium for various reference tether angles $\beta$. The bar plot displays the mean and one standard deviation on each side. \changed{The real-world power data is consistently higher than predicted by the first-principle power analysis in Section \ref{sec:pow-analysis} by roughly $10 \%$. However, the prediction is within the standard deviation of the averaged measurement and we are still able to observe trends aligning with the analytical predictions. The power consumption increases as $\beta$ increases for the static case, while we observe a flat trend in the rotating case when $\omega_C = \omega_C^*$.} For low $\beta$ values the standard deviation bars from the two cases intersect, but for higher $\beta$ values the separation widens, \changed{showing a statistically significant decrease in the average power consumption for the rotating configuration, with a $18.4\%$ decrease between the mean values at $\beta = 60^\circ$.}

There are several external factors not considered in our analysis that likely affect experimental data. The consistent \changed{10\% increase} over the predicted value could be attributed to feedback control actions, such as additional differential thrust needed to account for a small center-of-mass attachment point error. In the rotating case, the quadcopters must also maintain their yaw-rate at $\omega_C$, which requires additional differential thrust to oppose the resulting drag. \changed{Another source of error, particularly for the rotating case, is the assumption that the propeller Figure of Merit measured on the thrust stand is consistent for non-static cases. This could explain why the gap between the rotating and static case is smaller in the experimental results compared to what was predicted using the power model. Battery effects could also contribute to higher power measurements than expected, as larger voltage sag under the increased load of carrying the payload would increase the current running through the powertrain, potentially reducing efficiency.} Regardless, our model captures the patterns we observe in the experimental findings, providing confidence in using this approach as a design tool for larger scale systems in applications such as construction, logistics and transportation.

\begin{figure}
    \centering
    \includegraphics[width=0.8\linewidth]{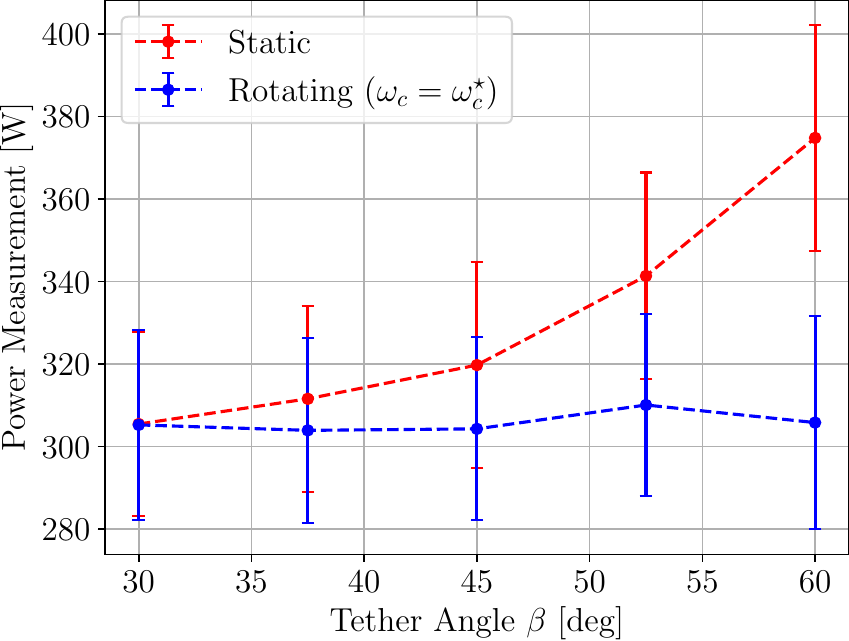}
    \caption{Experimental comparison of power consumption as a function of reference tether angle $\beta$ in static and rotating conditions. The power mean and standard deviations are computed from $20\,\mathrm{s}$ windows of onboard current and voltage readings. \changed{At $\beta = 60^\circ$, the rotating configuration decreases average power consumption by $18.4 \%$ compared to the static configuration.}}
    \label{fig:exp-results}
\end{figure}




\subsection{\changed{Outdoor Demonstration}}

\changed{The control architecture for the rotating configuration is demonstrated outdoors using fully onboard estimation. The payload and each quadcopter run individual onboard six degree of freedom estimators, which is sent to the onboard Jetson Orin Nano to be published on the ROS network. The outer-loop, system level controller runs on a laptop hosting the ROS network, which subscribes to the individual 6DOF estimates.}

\begin{figure}[t!]
    \centering
    \includegraphics[width=0.8\linewidth]{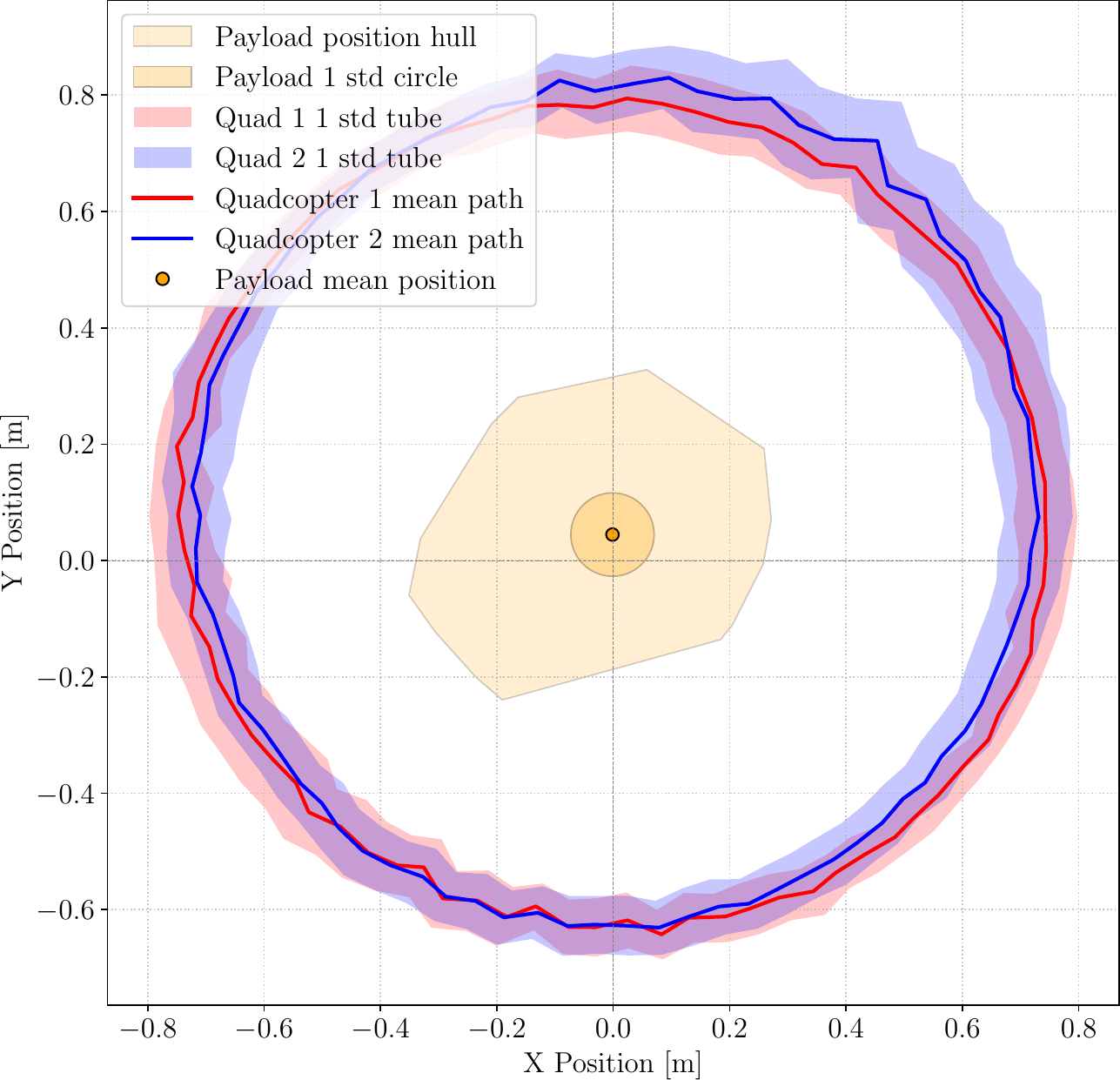}
    \caption{\changed{Mean quadcopter and payload positions in ENU frame during a $30$ second outdoor hover experiment with a desired payload position at the origin and tether angle $\beta = 45^\circ$. In yellow is the convex hull and one standard deviation of the payload position. In red and blue are the two mean paths of the quadcopters relative to the origin, as well as their one standard deviation distance away from the origin.} }
    \label{fig:outdoortracking}
\end{figure}

\changed{The tracking results of an outdoor hover experiment is shown in Fig. \ref{fig:outdoortracking}. This demonstrates sufficient payload position tracking performance in outdoor scenarios, with wind disturbances, sensor noise, and other outdoor environmental factors. Video of the outdoor experiment can be found in the attached supplementary video.}

\section{Conclusion and Future Work}
In summary, this work proposed a novel perspective on tether-suspended aerial transport using quadcopters. The key insight is that rotation about a vertical axis enables the vehicles to produce purely vertical thrust as the horizontal tether tension is balanced by centrifugal acceleration, reducing the power consumption of the transport system.
We conducted a first-principle analysis to predict the efficiency gains achieved by the system rotation, and experimentally validated our method through indoor flights confirming the first-principle predictions. Our experimental findings show that power savings of up to $\changed{18.4\%}$ are achievable by having the system operate at a rotating equilibrium. Our results demonstrate the potential of the proposed strategy to extend flight endurance in cooperative aerial transportation tasks and make such systems more practically relevant. 

Future work includes analysis on how robustness to external perturbations is affected by the rotating conditions, as well as other modes of flight. \changed{The analysis presented also assumed an equally spaced configuration to generalize to a greater number of vehicles, but further work can analyze how different vehicle layouts affect robustness and performance.} Our approach naturally extends to fixed-wing aircraft, which could be used for vertical take-off and landing transportation of payloads using our method, potentially reducing the power consumption even more significantly. \changed{The current implementation relies on the payload receiving direct measurements and running its own estimator, but future work could integrate a system wide estimator that does not rely on sensors mounted on the payload.} \changed{Additionally, the point-mass payload assumption is strong and required the payload tether attachment to be above its center of mass to remove uncontrolled oscillations of the payload. Future work could model the payload as a rigid body to capture and control these rotational dynamics. Multiple payload tether attachment points could be considered as well, but we believe a single attachment point analysis is still valuable as a common method in practice.}

\section*{Acknowledgment}
The authors gratefully acknowledge financial support from Rothberg Catalyzer, Mitsubishi HC Capital America, and The David Bogy Graduate Fellowship Fund.
The experimental testbed at the \mbox{HiPeRLab} is the result of contributions of many people, a full list of which can be found at \href{https://hiperlab.berkeley.edu/members/}{\texttt{hiperlab.berkeley.edu/members/}}.

\bibliographystyle{ieeetr}
\balance
\bibliography{refs}

@article{villa2020survey,
  title={A survey on load transportation using multirotor UAVs},
  author={Villa, Daniel KD and Brandao, Alexandre S and Sarcinelli-Filho, M{\'a}rio},
  journal={Journal of Intelligent \& Robotic Systems},
  volume={98},
  pages={267--296},
  year={2020},
  publisher={Springer}
}

@article{estevez2024review,
  title={Review of aerial transportation of suspended-cable payloads with quadrotors},
  author={Estevez, Julian and Garate, Gorka and Lopez-Guede, Jose Manuel and Larrea, Mikel},
  journal={Drones},
  volume={8},
  number={2},
  pages={35},
  year={2024},
  publisher={MDPI}
}

@ARTICLE{zhang2023formation,
  author={Zhang, Xiaozhen and Zhang, Fan and Huang, Panfeng},
  journal={IEEE Transactions on Automation Science and Engineering}, 
  title={Formation Planning for Tethered Multirotor UAV Cooperative Transportation With Unknown Payload and Cable Length}, 
  year={2024},
  volume={21},
  number={3},
  pages={3449-3460},
  doi={10.1109/TASE.2023.3279827}}

@article{sun2025agile,
  title={Agile and cooperative aerial manipulation of a cable-suspended load},
  author={Sun, Sihao and Wang, Xuerui and Sanalitro, Dario and Franchi, Antonio and Tognon, Marco and Alonso-Mora, Javier},
  journal={Science Robotics},
  volume={10},
  number={107},
  pages={eadu8015},
  year={2025},
  publisher={American Association for the Advancement of Science}
}

@Inbook{mellinger2013cooperative,
author="Mellinger, Daniel
and Shomin, Michael
and Michael, Nathan
and Kumar, Vijay",
editor="Martinoli, Alcherio
and Mondada, Francesco
and Correll, Nikolaus
and Mermoud, Gr{\'e}gory
and Egerstedt, Magnus
and Hsieh, M. Ani
and Parker, Lynne E.
and St{\o}y, Kasper",
title="Cooperative Grasping and Transport Using Multiple Quadrotors",
bookTitle="Distributed Autonomous Robotic Systems: The 10th International Symposium",
year="2013",
publisher="Springer Berlin Heidelberg",
address="Berlin, Heidelberg",
pages="545--558",
isbn="978-3-642-32723-0",
doi="10.1007/978-3-642-32723-0_39",
url="https://doi.org/10.1007/978-3-642-32723-0_39"
}

@ARTICLE{bosio2025automated,
  author={Bosio, Carlo and Mueller, Mark W.},
  journal={IEEE Robotics and Automation Letters}, 
  title={Automated Layout and Control Co-Design of Robust Multi-UAV Transportation Systems}, 
  year={2025},
  volume={10},
  number={4},
  pages={3956-3963},
  doi={10.1109/LRA.2025.3547307}}

@inproceedings{oishi2021autonomous,
  title={Autonomous cooperative transportation system involving multi-aerial robots with variable attachment mechanism},
  author={Oishi, Koshi and Jimbo, Tomohiko},
  booktitle={2021 IEEE/RSJ International Conference on Intelligent Robots and Systems (IROS)},
  pages={6322--6328},
  year={2021},
  organization={IEEE}
}

@inproceedings{jain2020flying,
  title={Flying batteries: In-flight battery switching to increase multirotor flight time},
  author={Jain, Karan P and Mueller, Mark W},
  booktitle={2020 IEEE International Conference on Robotics and Automation (ICRA)},
  pages={3510--3516},
  year={2020},
  organization={IEEE}
}

@article{jackson2020scalable,
  title={Scalable cooperative transport of cable-suspended loads with UAVs using distributed trajectory optimization},
  author={Jackson, Brian E and Howell, Taylor A and Shah, Kunal and Schwager, Mac and Manchester, Zachary},
  journal={IEEE Robotics and Automation Letters},
  volume={5},
  number={2},
  pages={3368--3374},
  year={2020},
  publisher={IEEE}
}

@inproceedings{ritz2013carrying,
  title={Carrying a flexible payload with multiple flying vehicles},
  author={Ritz, Robin and D'Andrea, Raffaello},
  booktitle={2013 IEEE/RSJ International Conference on Intelligent Robots and Systems},
  pages={3465--3471},
  year={2013},
  organization={IEEE}
}

@inproceedings{schulz2015high,
  title={High-speed, steady flight with a quadrocopter in a confined environment using a tether},
  author={Schulz, Maximilian and Augugliaro, Federico and Ritz, Robin and D'Andrea, Raffaello},
  booktitle={2015 IEEE/RSJ International Conference on Intelligent Robots and Systems (IROS)},
  pages={1279--1284},
  year={2015},
  organization={IEEE}
}

@article{Williams2007DynamicsPart1,
  author  = {Williams, Paul and Trivailo, Pavel},
  title   = {{Dynamics of Circularly Towed Aerial Cable Systems, Part 1: Optimal Equilibrium Configurations and Their Stability}},
  journal = {Journal of Guidance, Control, and Dynamics},
  volume  = {30},
  number  = {3},
  pages   = {753--765},
  year    = {2007},
  month   = {May},
  doi     = {10.2514/1.22906},
  url     = {https://www.researchgate.net/publication/238189161_Dynamics_of_Circularly_Towed_Cable_Systems_Part_1_Optimal_Configurations_and_Their_Stability}
}

@article{Williams2012Multi,
author = {Williams, Paul and Ockels, Wubbo},
title = {Dynamics of Towed Payload System Using Multiple Fixed-Wing Aircraft},
journal = {Journal of Guidance, Control, and Dynamics},
volume = {32},
number = {6},
pages = {1766-1780},
year = {2009},
doi = {10.2514/1.44371},
URL = { 
        https://doi.org/10.2514/1.44371
},
eprint = { 
        https://doi.org/10.2514/1.44371
}
}

@inproceedings{WilliamsTether,
  author    = {Paul Williams and Peter Lapthorne and Pavel Trivailo},
  title     = {{Circularly-Towed Lumped Mass Cable Model Validation from Experimental Data}},
  booktitle = {AIAA Modeling and Simulation Technologies Conference and Exhibit},
  year      = {2006},
  month     = {August},
  publisher = {American Institute of Aeronautics and Astronautics},
  address   = {Keystone, Colorado},
  doi       = {10.2514/6.2006-6817},
  url       = {https://arc.aiaa.org/doi/abs/10.2514/6.2006-6817}
}

@inproceedings{IndoorExperiment,
author = {Chapdelaine, Bruno and Rancourt, David and Ledoux, Guillaume},
year = {2019},
month = {05},
pages = {1-10},
title = {Experimental Validation of Vertical Lifting Capabilities of Circling Tethered Fixed Wing UAVs},
doi = {10.4050/F-0075-2019-14482}
}

@inproceedings{Quenneville2023Experimental,
  author    = {Quenneville, Samuel and Th{\'e}rien, Francis and Verrette, Jessy and Lucking Bigu{\'e}, Jean-Philippe and Bouchard, Mathieu and Doguet, Maxime and Rancourt, David},
  title     = {{Experimental Demonstration of the Lifting Capability of a Towed Payload Using Multiple Fixed-wing UAVs}},
  booktitle = {Vertical Flight Society 79th Annual Forum \& Technology Display},
  year      = {2023},
  month     = {May},
  location  = {West Palm Beach, Florida},
  url       = {https://www.researchgate.net/publication/371383180_Experimental_Demonstration_of_the_Lifting_Capability_of_a_Towed_Payload_Using_Multiple_Fixed-wing_UAVs}
}

@article{lahmeri2022charging,
  title={Charging techniques for UAV-assisted data collection: Is laser power beaming the answer?},
  author={Lahmeri, Mohamed-Amine and Kishk, Mustafa A and Alouini, Mohamed-Slim},
  journal={IEEE Communications Magazine},
  volume={60},
  number={5},
  pages={50--56},
  year={2022},
  publisher={IEEE}
}

@article{alyassi2022autonomous,
  title={Autonomous recharging and flight mission planning for battery-operated autonomous drones},
  author={Alyassi, Rashid and Khonji, Majid and Karapetyan, Areg and Chau, Sid Chi-Kin and Elbassioni, Khaled and Tseng, Chien-Ming},
  journal={IEEE Transactions on Automation Science and Engineering},
  volume={20},
  number={2},
  pages={1034--1046},
  year={2022},
  publisher={IEEE}
}

@article{mueller2016relaxed,
  title={Relaxed hover solutions for multicopters: Application to algorithmic redundancy and novel vehicles},
  author={Mueller, Mark W and D’Andrea, Raffaello},
  journal={The International Journal of Robotics Research},
  volume={35},
  number={8},
  pages={873--889},
  year={2016},
  publisher={SAGE Publications Sage UK: London, England}
}

@book{helicopteraero,
    author = {J.G. Leishman},
    title = {Principles of Helicopter Aerodynamics},
    edition = {2},
    publisher = {Cambridge University Press},
    year = {2000} 
}

@article{experimentallevelflightquadcopters,
author = {Bart Theys and Joris De Schutter},
title ={Forward flight tests of a quadcopter unmanned aerial vehicle with various spherical body diameters},

journal = {International Journal of Micro Air Vehicles},
volume = {12},
number = {},
pages = {1756829320923565},
year = {2020},
doi = {10.1177/1756829320923565},

URL = { 
    
        https://doi.org/10.1177/1756829320923565
    
    

},
eprint = { 
    
        https://doi.org/10.1177/1756829320923565
    
    

}
}

@Article{levelflightpower,
AUTHOR = {Hwang, Myeong-hwan and Cha, Hyun-Rok and Jung, Sung Yong},
TITLE = {Practical Endurance Estimation for Minimizing Energy Consumption of Multirotor Unmanned Aerial Vehicles},
JOURNAL = {Energies},
VOLUME = {11},
YEAR = {2018},
NUMBER = {9},
ARTICLE-NUMBER = {2221},
URL = {https://www.mdpi.com/1996-1073/11/9/2221},
ISSN = {1996-1073},
DOI = {10.3390/en11092221}
}

@ARTICLE{swarmpoweranalysis,
  author={Cabuk, Umut Can and Tosun, Mustafa and Dagdeviren, Orhan and Ozturk, Yusuf},
  journal={IEEE Transactions on Intelligent Transportation Systems}, 
  title={Modeling Energy Consumption of Small Drones for Swarm Missions}, 
  year={2024},
  volume={25},
  number={8},
  pages={10176-10189},
  doi={10.1109/TITS.2024.3350042}}


\end{document}